\documentclass{article}

\usepackage{microtype}
\usepackage{graphicx}
\usepackage{subcaption}
\usepackage{caption} %
\usepackage{booktabs} %
\usepackage{multirow}
\usepackage{xcolor} %

\usepackage{algorithm}

\usepackage{hyperref}

\usepackage{enumitem}                            %

\usepackage[preprint]{icml2026}

\usepackage{amsmath}
\usepackage{amssymb}
\usepackage{mathtools}
\usepackage{amsthm}

\usepackage[capitalize,noabbrev]{cleveref}

\theoremstyle{plain}

\theoremstyle{definition}

\theoremstyle{remark}

\usepackage[textsize=tiny]{todonotes}

\begin{document}

\twocolumn[
  \icmltitle{Geometry-Aware Rotary Position Embedding for Consistent Video World Model}

  \icmlsetsymbol{equal}{*}

  \begin{icmlauthorlist}
    \icmlauthor{Chendong Xiang}{equal,thu}
    \icmlauthor{Jiajun Liu}{equal,ruc}
    \icmlauthor{Jintao Zhang}{equal,thu}
    \icmlauthor{Xiao Yang}{thu}
    \icmlauthor{Zhengwei Fang}{thu}
    \icmlauthor{Shizun Wang}{nus}
    \icmlauthor{Zijun Wang}{pku}
    \icmlauthor{Yingtian Zou}{sjtu}
    \icmlauthor{Hang Su}{thu}
    \icmlauthor{Jun Zhu}{thu}
  \end{icmlauthorlist}

  \icmlaffiliation{thu}{Dept. of Comp. Sci. and Tech., Institute for AI, BNRist Center, THBI Lab, Tsinghua-Bosch Joint ML Center, Tsinghua University}
  \icmlaffiliation{ruc}{Gaoling School of Artificial Intelligence, Renmin University of China}
  \icmlaffiliation{nus}{National University of Singapore}
  \icmlaffiliation{pku}{Peking University}
  \icmlaffiliation{sjtu}{School of Artificial Intelligence, Shanghai Jiao Tong University}

  \icmlcorrespondingauthor{Hang Su}{suhangss@tsinghua.edu.cn}
  \icmlcorrespondingauthor{Jun Zhu}{dcszj@tsinghua.edu.cn}

  \vskip 0.2in

  \centering
  {\centering\includegraphics[width=0.95\textwidth]{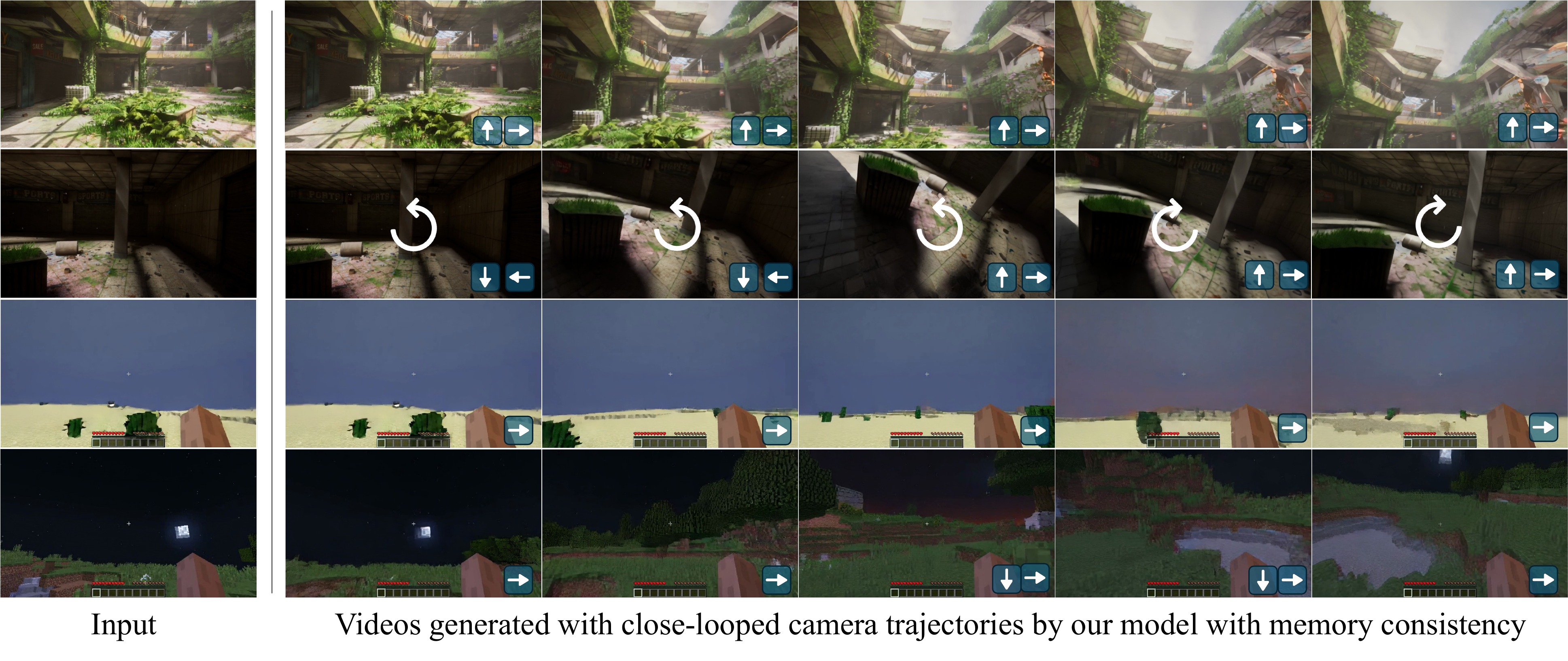}\par}
  \captionof{figure}{\textbf{Camera-controlled video generation with ViewRope.} Up: generated video with camera trajectories with loop closure (rotate-away-rotate-back). 
Down: generated videos with high motion minecraft gaming. ViewRope maintains consistent scene appearance when the camera revisits previously observed viewpoints.}
  \label{fig:teaser}
  \vskip 0.2in
]

\printAffiliationsAndNotice{\icmlEqualContribution}

\begin{abstract}
    Predictive world models that simulate future observations under explicit camera control are fundamental to interactive AI. Despite rapid advances, current systems lack spatial persistence: they fail to maintain stable scene structures over long trajectories, frequently hallucinating details when cameras revisit previously observed locations. We identify that this geometric drift stems from reliance on screen-space positional embeddings, which conflict with the projective geometry required for 3D consistency. We introduce \textbf{ViewRope}, a geometry-aware encoding that injects camera-ray directions directly into video transformer self-attention layers. By parameterizing attention with relative ray geometry rather than pixel locality, ViewRope provides a model-native inductive bias for retrieving 3D-consistent content across temporal gaps. We further propose \textbf{Geometry-Aware Frame-Sparse Attention}, which exploits these geometric cues to selectively attend to relevant historical frames,  improving efficiency without sacrificing memory consistency. We also present \textbf{ViewBench}, a diagnostic suite measuring loop-closure fidelity and geometric drift. Our results demonstrate that ViewRope substantially improves long-term consistency while reducing computational costs.
\end{abstract}

\section{Introduction}
\label{sec:intro}

Developing pose-conditioned visual world models---predictive simulators that generate future observations under
an \emph{explicit viewpoint trajectory}---is fundamental to interactive AI systems~\citep{ha2018world, hafner2023mastering, yang2025longlive, zhang2025matrixgameinteractiveworldfoundation}.
Despite remarkable progress in open-domain video diffusion models
\citep{wan2025wanopenadvancedlargescale, kong2025hunyuanvideosystematicframeworklarge, veo3techreport2025, bao2024viduhighlyconsistentdynamic},
current generators fail to maintain \emph{long-term geometric consistency}: as viewpoints evolve, they do not preserve stable scene structures that can be revisited.
This deficiency becomes most apparent in \emph{loop-closure} trajectories (e.g., \emph{rotate-away-rotate-back}), where the camera returns to a previously
observed viewpoint after traversing elsewhere.
A consistent world model should reconstruct identical structures and appearances upon revisiting. Instead, existing generators frequently hallucinate new details or
drift, revealing the absence of a reliable mechanism for retaining and retrieving 3D-consistent content over time.

Existing approaches typically address this challenge through two strategies.
The first enlarges context via external retrieval or memory---selecting historical frames 
based on field-of-view overlap or maintaining explicit 3D spatial 
structures~\citep{yu2025contextmemorysceneconsistentinteractive, wu2025video, 
huang2025memoryforcingspatiotemporalmemory, oshima2025worldpack}.
However, these mechanisms rely on pixel-level concatenation or 
external data structures rather than being integrated into the model's internal 
representation natively; this often incurs substantial compute and can become 
brittle when histories are long or camera motion is complex.
The second employs geometry-first pipelines such as
3D Gaussian Splatting~\citep{kerbl20233dgaussiansplattingrealtime} and specialized 
novel-view synthesis transformers~\citep{jin2024lvsm, gao2024cat3d, 
li2025camerasrelativepositionalencoding}---which enforce strict 3D consistency 
but typically sacrifice open-domain flexibility.

We trace this failure to a bottleneck in \emph{positional modeling}.
Most video transformers parameterize
space--time structure in \emph{screen coordinates} $(x,y,t)$ via learned absolute/relative embeddings~\citep{su2023roformerenhancedtransformerrotary}.
However, under camera rotation and translation, correspondence is dictated by projective geometry: the same 3D point can
map to widely separated image locations across time, and conversely, nearby pixels need not be co-visible. As a
result, screen-space positional bias is misaligned with the invariances required for view-consistent generation,
inducing \emph{geometric drift} that compounds over long trajectories and becomes most evident at loop closure when
the camera revisits previously observed viewpoints (Sec.~\ref{sec:experiments}). The central challenge is thus to
equip transformers with a mechanism to identify and reuse \emph{the same physical content} across temporally distant
tokens whose image-plane coordinates are decorrelated by camera motion—without resorting to explicit 3D scene
reconstruction or compromising open-domain generative flexibility.

Our key insight is that long-horizon view consistency is governed by \emph{angular correspondence of viewing
directions}, rather than locality in the image plane---and that this geometric prior can be embedded 
\emph{directly into the attention mechanism} without external memory structures. Under calibrated camera motion, two tokens are likely to be
informative to each other when their associated camera rays are \emph{co-visible} (i.e., intersect the same physical
content), even if they are separated by long temporal gaps and occupy unrelated pixel coordinates. Motivated by this
principle, we propose \textbf{ViewRope}, a geometry-aware positional encoding that injects patch-level viewing-ray
directions into self-attention through ray-based rotary transformations of the query/key features. In contrast to
standard 2D/3D RoPE, which encodes pixel-space offsets, ViewRope parameterizes attention as a function of \emph{relative
ray geometry}, yielding a model-native inductive bias for retrieving and reusing consistent 3D content from long
histories. 
Unlike explicit memory approaches that maintain external data structures, ViewRope realizes geometric correspondence 
\emph{implicitly} through attention itself, offering a lightweight and complementary mechanism.
To further support long-context generation, we introduce \textbf{Geometry-Aware Frame-Sparse Attention},
which leverages geometry-conditioned relevance to select a small set of co-visible historical frames, replacing
quadratic dense attention with geometry-driven sparsity while preserving loop-closure fidelity.

To evaluate view consistency directly, we introduce \textbf{ViewBench}, a diagnostic benchmark tailored to
camera-conditioned long-horizon generation. Unlike generic perceptual metrics (e.g., FVD/IS) that primarily measure
frame quality, ViewBench targets loop-closure trajectories such as \emph{rotate-away--rotate-back} and quantifies
revisit fidelity and geometric drift. Experiments show that our approach substantially improves view-consistency on
ViewBench while remaining efficient, narrowing the gap between geometry-rigid 3D pipelines and open-domain diffusion
generators. Our contributions are summarized as follows:
\begin{itemize}[leftmargin=*, itemsep=2pt, topsep=2pt]
    \item \textbf{ViewRope}: a geometric positional encoding that injects \emph{patch-level} camera-ray directions into
    attention, yielding a model-native inductive bias for long-term geometric consistency.
    \item \textbf{Geometry-Aware Frame-Sparse Attention}: an efficient, geometry-conditioned retrieval mechanism that
    selects co-visible historical frames, enabling consistent long-video generation with low latency.
    \item \textbf{ViewBench}: a targeted evaluation suite for quantifying view-consistency and loop-closure behavior in
    camera-conditioned video generation models.
\end{itemize}

\section{Related Work}

\subsection{Conditioning Transformers on Camera Geometry}
Camera conditioning is essential for binding geometric viewpoint information to visual tokens in multiview and video transformers.
A dominant approach is to encode camera parameters into raymaps---per-pixel 6D embeddings containing ray origins and directions or Plücker coordinates~\cite{mildenhall2021nerf,zhang2024cameras,gao2024cat3dcreate3dmultiview,jin2024lvsm,weber2025fillerbuster}.
Concatenating these raymaps to input tokens allows models to condition on both intrinsics and extrinsics~\cite{zhang2024cameras}.
However, raymaps typically rely on a global reference frame, which can be arbitrary and hinder generalization~\cite{mildenhall2019local,guizilini2025zero}.

To mitigate this, recent methods employ relative attention-level encodings that leverage the geometric relationship between views.
CaPE and GTA embed relative SE(3) pose by applying transformations directly to attention mechanisms~\cite{safin2023repast,miyato2024gtageometryawareattentionmechanism,kong2024eschernetgenerativemodelscalable}.
PRoPE~\cite{li2025camerasrelativepositionalencoding} models the complete relative projective transformation, encoding both camera intrinsics and extrinsics within the attention mechanism to better ground visual tokens in 3D space.
While these methods improve view synthesis and avoid global frame dependency, they assume shared camera poses for all pixels within a view, lacking fine-grained geometric modeling.

\subsection{Interactive World Models}
Recent interactive world models enable controllable simulation of physical environments by conditioning video generation on user actions and historical context \cite{che2024gamegenxinteractiveopenworldgame,zhang2025matrixgameinteractiveworldfoundation,li2025hunyuangamecrafthighdynamicinteractivegame}.
To support real-time interaction, the field has seen a paradigm shift from bidirectional diffusion to causal or autoregressive architectures \cite{valevski2024diffusion, yin2025slowbidirectionalfastautoregressive,yang2025longliverealtimeinteractivelong,huang2025selfforcingbridgingtraintest}, often utilizing KV-caching and distillation to accelerate inference.
While scaling training on massive gameplay datasets enables foundation models like Matrix-Game~\citep{zhang2025matrixgameinteractiveworldfoundation} and Hunyuan-GameCraft \cite{li2025hunyuangamecrafthighdynamicinteractivegame} to achieve high-dynamic controllability, maintaining long-term spatial consistency---particularly during scene revisits---remains a critical bottleneck~\citep{lian2025memoryaidedworldmodelsbenchmarking}.

To address this limitation, recent works introduce explicit memory mechanisms.
Context-as-Memory~\citep{yu2025contextmemorysceneconsistentinteractive} retrieves historical frames based on field-of-view overlap and concatenates them into the generation context.
Memory Forcing~\citep{huang2025memoryforcingspatiotemporalmemory} incorporates a geometry-indexed spatial memory to enforce coherence across extended horizons.
More ambitiously, Wu et al.~\citep{wu2025video} propose augmenting world models with explicit 3D point-cloud memory inspired by human spatial cognition, while WorldPack~\citep{oshima2025worldpack} compresses trajectory history via packing and selective retrieval.
These memory-based approaches demonstrate improved loop-closure consistency but rely on external data structures that are not native to the attention mechanism.
Our work instead embeds geometric correspondence \emph{directly} into positional encoding, enabling implicit memory retrieval through attention without auxiliary modules.

\subsection{Sparse Attention with Long Sequence}
The quadratic complexity of the self-attention mechanism with respect to sequence length poses a significant challenge for modeling long sequences. 
To address this bottleneck, recent works have explored sparse attention mechanisms that reduce computational cost by attending to only a subset of tokens. 
In both language and vision domains, various sparse attention methods have been proposed. These approaches typically rely on learnable schemes~\citep{gao2025seerattentionlearningintrinsicsparse, deepseekai2025deepseekv32pushingfrontieropen}, pattern-based selection~\citep{lai2025flexprefillcontextawaresparseattention, xi2025sparsevideogenacceleratingvideo}, or low-cost dynamic estimation~\citep{zhang2025vsafastervideodiffusion, zhang2025spargeattentionaccuratetrainingfreesparse, Zhu_2025, zhu2025sampleattentionnearlosslessaccelerationlong}. 
However, in the setting of Autoregressive Diffusion for video generation, sparse attention remains relatively underexplored. 
Current state-of-the-art approaches typically rely on Sliding Window Attention to handle long temporal sequences. 
For instance, LongLive~\citep{yang2025longliverealtimeinteractivelong} utilizes short window attention combined with frame sinks to maintain efficiency and consistency during real-time interactive long video generation.

\section{Method}

\subsection{Problem Formulation}
We study \emph{pose-conditioned video generation} as a visual world model. Given an initial observation
$\mathbf{x}_0$ (or a short context $\mathbf{x}_{\le 0}$) and a target camera trajectory, the model generates a future
video $\mathbf{x}_{1:T}$ that is consistent with the specified viewpoint evolution.

Let $\mathcal{C}_{1:T}:=\{(\mathbf{R}_t,\mathbf{P}_t,\mathbf{K}_t)\}_{t=1}^T$ denote the camera trajectory, where
$\mathbf{R}_t\in SO(3)$, $\mathbf{P}_t\in\mathbb{R}^3$, and $\mathbf{K}_t\in\mathbb{R}^{3\times 3}$ represent camera rotation, translation, and intrinsics at time $t$, respectively. Optionally, we map the trajectory to a low-level \emph{action prompt} (e.g., turn, move) and provide a global text description of the scene. We denote the resulting text/action conditioning by $\mathcal{Y}$. Our goal is to learn the conditional distribution
\begin{equation}
p_\theta(\mathbf{x}_{1:T}\mid \mathbf{x}_{\le 0}, \mathcal{C}_{1:T}, \mathcal{Y}).
\end{equation}

Standard video generators mainly enforce \emph{local temporal coherence}, which only constrains adjacent frames. For a generic photometric/perceptual distance $d(\cdot,\cdot)$, this is captured by
\begin{equation}
\mathcal{L}_{\text{temp}}(\theta)
:= \mathbb{E}_{\mathbf{x} \sim p_\theta}\!\left[\sum_{t=2}^{T} d\!\left(\mathbf{x}_t,\mathbf{x}_{t-1}\right)\right].
\end{equation}
Such objectives do not prevent \emph{long-horizon geometric drift} under camera motion, because screen-space proximity is not aligned with physical correspondence.

In contrast, a pose-conditioned world model must satisfy a \emph{loop-closure} requirement: if the camera at time $t$ revisits (approximately) a viewpoint observed at some $k\ll t$, then the rendered observations should agree up to projective geometry on their co-visible region. Concretely, define the revisit indicator
\begin{equation}
w_{t,k} := \mathbb{I}\!\left(\Delta(\mathcal{C}_t,\mathcal{C}_k)\le \varepsilon\right),\qquad k<t,
\end{equation}
where $\Delta(\cdot,\cdot)$ measures pose similarity (e.g., rotation/ translation discrepancy under calibrated intrinsics) and $\varepsilon$ is a tolerance threshold.
When $w_{t,k}=1$, the generated frames $\mathbf{x}_t$ and $\mathbf{x}_k$ should be photometrically consistent on their co-visible region $\Omega_{t,k}$. Let $\mathcal{W}_{k\leftarrow t}$ denote the projective warp from $t$ to $k$ (see Appendix~\ref{app:loop_closure_formulation} for details). We formalize this via a loop-closure loss:
\begin{equation}
\begin{split}
\mathcal{L}_{\text{lc}}(\theta)
&:=
\mathbb{E}_{\mathbf{x} \sim p_\theta}\!\bigg[
\sum_{t=1}^{T}\sum_{k<t} w_{t,k} \\
&\quad \cdot \sum_{\mathbf{u}\in\Omega_{t,k}}
\rho\!\left(
\mathbf{x}_t(\mathbf{u}) -
\mathbf{x}_k\!\left(\mathcal{W}_{k\leftarrow t}(\mathbf{u})\right)
\right)
\bigg],
\end{split}
\label{eq:loop_closure_loss}
\end{equation}
where $\rho(\cdot)$ is a robust penalty.

The difficulty is that this loop-closure objective couples frames across \emph{arbitrarily long} temporal gaps: achieving loop closure requires retrieving geometrically corresponding content from a long history under causal (streaming) generation.
Rather than explicitly optimizing a pixel-level consistency loss, we parameterize $p_\theta$ with a Diffusion Transformer (DiT) and inject 3D view geometry into its attention mechanism, so that (i) attention scores become sensitive to \emph{relative viewing rays},
and (ii) the model can efficiently select and attend to the most geometrically relevant historical frames while
generating online.

\begin{figure*}[t]
    \centering
    \includegraphics[width=0.95\textwidth]{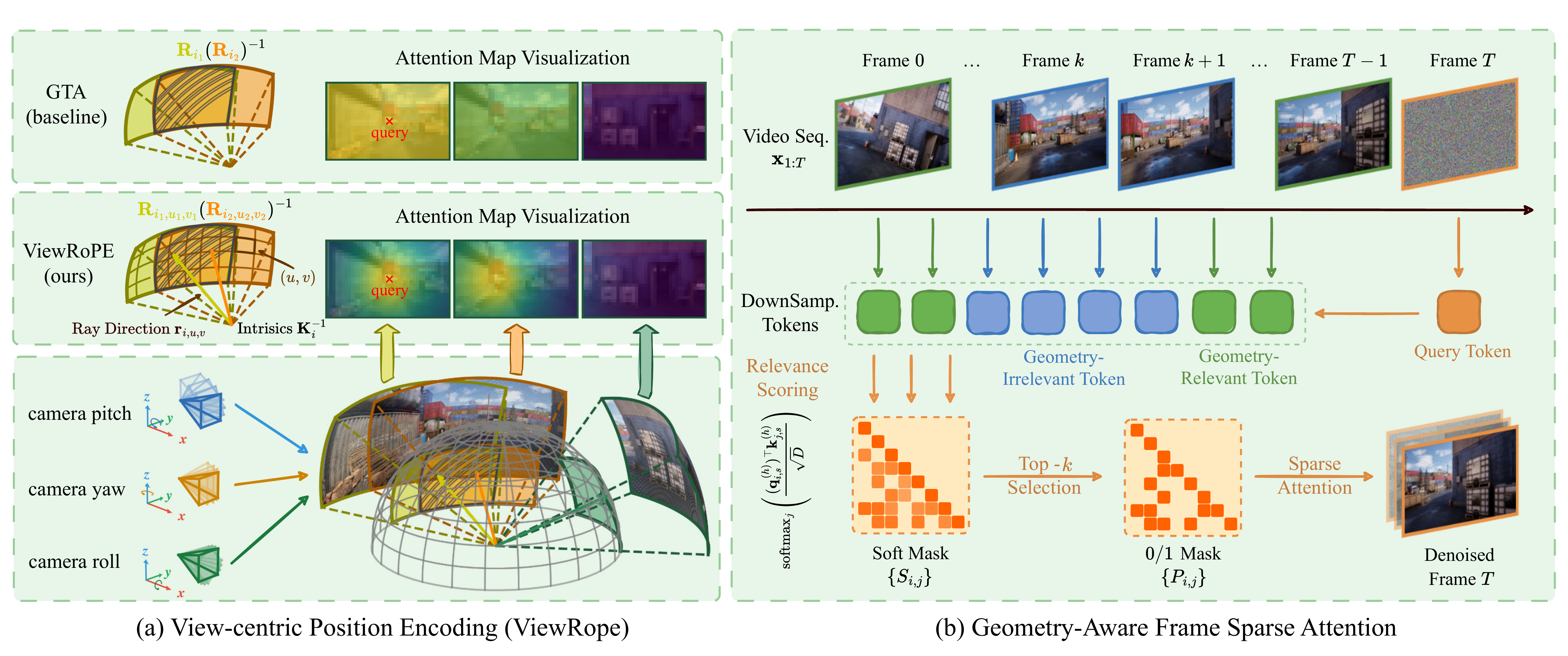}
    \vspace{-3mm}
    \caption{\textbf{Method overview.}
    \textbf{(a) ViewRope} computes per-patch viewing rays from intrinsics, constructs local rotations, and rotates
    query/key feature subvectors in attention. The resulting dot product encodes relative angular relationships
    between viewing rays.
    \textbf{(b) Geometry-Aware Frame Sparse Attention} estimates block (frame) relevance and selects top-$k$
    geometrically relevant historical frames, replacing quadratic dense attention with geometry-driven sparsity.}
    \label{fig:method_overview}
    \vspace{-4mm}
\end{figure*}

\subsection{ViewRope: View-Centric Positional Encoding in Attention}
\label{sec:viewrope}
We introduce \emph{View-centric Position Encoding} (\textbf{ViewRope}), encoding each token's 3D viewing
direction \emph{directly} into the attention mechanism. Unlike 2D/3D positional embeddings or global pose tokens,
ViewRope assigns a \emph{per-patch} rotation derived from camera intrinsics/extrinsics, so attention can operate on
relative view geometry at patch granularity.

\paragraph{Per-patch ray construction.}
For a patch centered at pixel coordinates $(u,v)$ in camera/view $i$, we compute its normalized viewing ray
$\mathbf{r}_{i,u,v}\in\mathbb{S}^2$ (the unit sphere) in the camera coordinate system using intrinsics $\mathbf{K}_i$:
\begin{equation}
\mathbf{r}_{i,u,v}
= \frac{\mathbf{K}_i^{-1}[u,v,1]^\top}{\|\mathbf{K}_i^{-1}[u,v,1]^\top\|_2}.
\end{equation}
We then build a local rotation $\mathbf{R}^{\text{local}}_{i,u,v}\in SO(3)$ that maps the canonical optical axis
$\mathbf{z}=[0,0,1]^\top$ to $\mathbf{r}_{i,u,v}$. Combining with the camera extrinsic rotation
$\mathbf{R}^{\text{cam}}_i$, we obtain a world-aligned view rotation:
\begin{equation}
\mathbf{R}_{i,u,v}=\mathbf{R}^{\text{cam}}_i \, \mathbf{R}^{\text{local}}_{i,u,v}.
\end{equation}

\paragraph{Rotating query/key feature subvectors.}
Let $\mathbf{q}\in\mathbb{R}^d$ and $\mathbf{k}\in\mathbb{R}^d$ be query/key vectors for a token.
We reserve a subset of channels that can be grouped into $m$ 3D subvectors (so $3m\le d$), and rotate each subvector
by $\mathbf{R}_{i,u,v}$:
\begin{equation}
\mathrm{VR}(\mathbf{q},\mathbf{R}_{i,u,v})=\mathbf{q}',
\end{equation}
where $
\mathbf{q}'_{3\ell:3\ell+3}=\mathbf{R}_{i,u,v}\mathbf{q}_{3\ell:3\ell+3},\ \ell=0,\dots,m-1$.
We apply the same transformation to keys. Intuitively, this aligns a portion of the feature space with the physical
viewing direction of each patch in world coordinates.

\paragraph{Geometry-aware attention scores.}
Consider a query token from view $i$ at $(u_i,v_i)$ and a key token from view $j$ at $(u_j,v_j)$.
Their rotated dot product becomes
\begin{equation}
\begin{split}
&\left\langle \mathrm{VR}(\mathbf{q},\mathbf{R}_{i,u_i,v_i}),\ \mathrm{VR}(\mathbf{k},\mathbf{R}_{j,u_j,v_j}) \right\rangle
\\&= \mathbf{q}^\top \mathbf{R}_{i,u_i,v_i}^\top \mathbf{R}_{j,u_j,v_j}\mathbf{k} 
= \mathbf{q}^\top \left(\mathbf{R}_{i,u_i,v_i}^{-1}\mathbf{R}_{j,u_j,v_j}\right)\mathbf{k}.
\end{split}
\end{equation}
The relative rotation $\mathbf{R}_{i,u_i,v_i}^{-1}\mathbf{R}_{j,u_j,v_j}$ captures the angular relationship between
the two viewing rays. This makes attention naturally sensitive to 3D view similarity, improving long-range recall and
loop closure consistency.

\subsection{Geometry-Aware Frame Sparse Attention}
\label{sec:sparse_attn}
Long-context generation with dense attention scales quadratically with sequence length. To support streaming
world-modeling over many frames, we adopt a \emph{frame-aligned block-sparse} attention scheme inspired by
SampleAttention~\citep{zhu2025sampleattentionnearlosslessaccelerationlong}. The key design is to set the block size
$B$ to exactly match one latent frame, so blocks correspond to time steps.
As ViewRope encodes 3D viewing geometry in the attention space, we can directly estimate frame-level geometric relevance.

Let $\mathbf{Q},\mathbf{K},\mathbf{V}\in\mathbb{R}^{L\times D}$ be token sequences, partitioned into $N$ blocks of
size $B$ ($L=NB$). Denote block $i$ as $\mathbf{Q}_i\in\mathbb{R}^{B\times D}$ and block $j$ as
$\mathbf{K}_j,\mathbf{V}_j\in\mathbb{R}^{B\times D}$.

\paragraph{Block relevance estimation (stochastic).}
Instead of computing full block-to-block attention, we sample a small set of token indices
$\mathcal{S}\subset\{1,\dots,B\}$ of size $K_s$ (e.g., $K_s=10$) and estimate a head-averaged affinity:
\begin{equation}
\tilde{S}_{ij}
= \frac{1}{HK_s}\sum_{h=1}^{H}\sum_{s\in\mathcal{S}}
\frac{(\mathbf{q}^{(h)}_{i,s})^\top \mathbf{k}^{(h)}_{j,s}}{\sqrt{D}}.
\label{eq:block_affinity}
\end{equation}
We then apply a causal constraint via a block mask $\mathbf{M}^{\text{causal}}$ (disallowing $j>i$ in the streaming
setting) by setting $\tilde{S}_{ij}=-\infty$ when $M^{\text{causal}}_{ij}=0$.

\paragraph{Top-$k$ block selection.}
For each query block $i$, we select the top-$k$ key blocks under $\tilde{S}_{ij}$ among valid past blocks:
\begin{equation}
\mathcal{T}_i = \mathrm{TopK}\big(\{\tilde{S}_{ij}\}_{j:\,M^{\text{causal}}_{ij}=1}\big).
\end{equation}
We always include $j=i$ to preserve local context. The final sparsity mask is
\begin{equation}
M_{ij} =
\begin{cases}
1 & \text{if } (j\in\mathcal{T}_i \ \text{or}\ j=i)\ \land\ M^{\text{causal}}_{ij}=1,\\
0 & \text{otherwise}.
\end{cases}
\end{equation}

\paragraph{Sparse attention computation.}
We compute attention for block $i$ only over selected blocks:
\begin{equation}
\mathbf{O}_i
=
\mathrm{softmax}\!\left(
\frac{\mathbf{Q}_i\,(\mathbf{K}_{\{j\,|\,M_{ij}=1\}})^\top}{\sqrt{D}}
\right)\mathbf{V}_{\{j\,|\,M_{ij}=1\}}.
\label{eq:sparse_output}
\end{equation}
With fixed $k$, the attention cost scales linearly with the number of frames, enabling efficient long-horizon
generation. We implement the sparse attention kernel with TileLang~\citep{wang2025tilelangcomposabletiledprogramming},
following the optimization principles of FlashAttention~\citep{dao2023flashattention}.

\paragraph{Training vs. inference (streaming with cache).}
We utilize teacher-forcing~\citep{arriola2025blockdiffusioninterpolatingautoregressive} to enbale AR-generation. During training, historical frames are taken from ground-truth latents, forming a clean KV cache;
the current denoising step queries this cache with ViewRope-rotated features and a top-$k$ frame mask.
During autoregressive inference, we maintain a KV cache of previously generated latent frames and apply the
same relevance estimation and top-$k$ selection at each denoising step, preserving causality while retrieving
geometrically relevant history. The detailed training and inference procedures are shown in Algorithm~\ref{alg:sparse_attn}.

\vspace{-10pt}
\subsection{Progressive Training Pipeline}
To stabilize adaptation to autoregressive streaming generation and long contexts, we employ a progressive schedule:

\textbf{Stage I: Short-clip teacher forcing.} Train on short clips (e.g., $\sim$17 frames) with teacher forcing to align the model with the autoregressive generation interface and caching behavior.
\textbf{Stage II: Introduce ViewRope.} Enable ViewRope while keeping clips short, allowing the model to learn view-conditioned correspondence without the confound of very long contexts.
\textbf{Stage III: Enable Frame Sparse Attention.} Activate frame-aligned block sparsity to adapt the model to long-context retrieval efficiently, while keeping sequence lengths moderate.
\textbf{Stage IV: Scale context length.} Increase training sequence length substantially under sparse attention, endowing the model with long-horizon video generation and improved loop-closure consistency.

\section{Experiments}
\label{sec:experiments}

We conduct experiments to validate the effectiveness of ViewRope. 
All experiments use the same backbone, training budget, and data to ensure fair comparison.

\subsection{Experimental Setup}
\label{subsec:setup}

\textbf{Datasets and Benchmarks.}
We evaluate on \textbf{ViewBench}, a diagnostic benchmark we construct to systematically evaluate view-consistency under camera motion.
Existing datasets hold limitations for expected and unified evaluation:
Context-as-Memory~\citep{yu2025contextmemorysceneconsistentinteractive} constrains camera motion to yaw-only rotation with pitch fixed at zero, preventing evaluation of general 3D rotations;
GF-Minecraft~\citep{yu2025gamefactorycreatingnewgames} focuses on action controllability without GT geometric overlap annotations required for attention analysis.
Table~\ref{tab:viewbench_comparison} summarizes these differences.
ViewBench addresses these gaps with:
\textbf{(i)} complete 3-axis rotation coverage (yaw, pitch, roll) with systematic angle sampling;
\textbf{(ii)} round-trip ``loop closure''~\citep{lian2025memoryaidedworldmodelsbenchmarking} trajectories where the camera returns to previously visited viewpoints;
\textbf{(iii)} 10 photorealistic UE5 environments spanning indoor, outdoor, urban, and natural settings.
The training set contains 1k+ video sequences (${\sim}$500k frames), and the evaluation set consists of 600 separately collected samples with non-overlapping trajectories. See Appendix~\ref{app:viewbench} for details.

\begin{table}[t]
\centering
\caption{\textbf{Comparison with existing datasets.} ViewBench fills gaps in evaluating view-consistency for interactive world models.}
\label{tab:viewbench_comparison}
\vspace{-2mm}
\small
\setlength{\tabcolsep}{4pt}  %
\begin{tabular}{l|ccc}
\toprule
\textbf{Property} & \textbf{CaM} & \textbf{GF-MC} & \textbf{ViewBench} \\
\midrule
Yaw & $\checkmark$ & $\checkmark$ & $\checkmark$ \\
Pitch & $\times$ & $\checkmark$ & $\checkmark$ \\
Roll & $\times$ & $\times$ & $\checkmark$ \\
Loop-closure traj. & $\times$ & $\times$ & $\checkmark$ \\
Controlled angles & $\times$ & $\times$ & $\checkmark$ \\
\bottomrule
\end{tabular}
\vspace{-3mm}
\end{table}

\textbf{Metrics.}
We report metrics across three categories: 1)\textbf{Visual Quality}: PSNR, SSIM, LPIPS---standard metrics for frame-level reconstruction quality; 2) \textbf{Loop Closure Error (LCE)}: LPIPS between the starting frame $\mathbf{x}_0$ and the generated frame $\mathbf{x}_t$ upon returning to the start pose, directly measuring persistent spatial memory. Lower is better.

\textbf{Baselines.}
We compare against two categories of methods:
1) \textbf{3D RoPE}~\citep{su2023roformerenhancedtransformerrotary}: Standard temporal-spatial RoPE without camera geometry, serving as a no-geometry baseline; 2) \textbf{GTA}~\citep{miyato2024gtageometryawareattentionmechanism}: Relative SE(3) transformation applied to attention, encoding extrinsics only.

Furthermore, we investigate the integration of ViewRope with various sparse attention mechanisms,
including our proposed Geometry-Aware Sparse Attention and sliding window attention,
to validate the effectiveness of our algorithm and demonstrate the efficiency improvements achieved through sparse attention patterns.

\textbf{Implementation Details.}
We build upon WAN 2.2 TI2V-5B~\citep{wan2025wanopenadvancedlargescale}, a text-and-image-to-video diffusion transformer, and adapt it for streaming video generation via teacher-forcing training.
The training data combines Context-as-Memory~\citep{yu2025contextmemorysceneconsistentinteractive}, GF-Minecraft~\citep{yu2025gamefactorycreatingnewgames}, and ViewBench at a 1:1:1 sampling ratio.
All RoPE variants are applied to the same channels for fair comparison.
The ViewBench evaluation set is \emph{separately collected} with non-overlapping trajectories.
See Appendix~\ref{app:implementation} for detailed training configurations.

\begin{table*}[t]
    \centering
    \caption{\textbf{Position encoding comparison on ViewBench.} We report visual quality (PSNR$\uparrow$, SSIM$\uparrow$, LPIPS$\downarrow$) and geometric consistency (LCE$\downarrow$) for 30\textdegree\ and 75\textdegree\ view synthesis. Best in \textbf{bold}.}
    \small
    \label{tab:main_results}
    \vspace{-2mm}
    \begin{tabular}{c|c c c c|c c c c}
        \toprule
        \multirow{2}{*}{\textbf{Method}} & \multicolumn{4}{c|}{\textbf{30 deg}} & \multicolumn{4}{c}{\textbf{75 deg}} \\
     & \textbf{PSNR} & \textbf{SSIM} & \textbf{LPIPS $\downarrow$} & \textbf{LCE $\downarrow$} & \textbf{PSNR} & \textbf{SSIM} & \textbf{LPIPS $\downarrow$} & \textbf{LCE $\downarrow$} \\
    \midrule
        3D RoPE & 17.09 & 0.4133 & 0.4219 & 0.4929 & 14.78 & 0.3634 & 0.5501 & 0.4831 \\
        GTA & 17.33 & 0.4325 & 0.4165 & 0.4707 & 15.12 & 0.3784 & 0.5403 & 0.4723 \\
        ViewRope (Ours) & \textbf{17.53} & \textbf{0.4378} & \textbf{0.4080} & \textbf{0.4497} & \textbf{15.27} & \textbf{0.3916} & \textbf{0.5398} & \textbf{0.4562} \\
        \hline
        \end{tabular}
    \vspace{-3mm}
    \end{table*}

\subsection{View Consistency Comparison}
\label{subsec:main}

Table~\ref{tab:main_results} presents the quantitative comparison on ViewBench.
We make three key observations:
\textbf{(1) ViewRope achieves the best loop closure performance.}
ViewRope reduces LCE by 4\% compared to GTA, the strongest baseline. 
This demonstrates that per-patch ray encoding provides more precise geometric alignment than per-camera projective encoding when revisiting previous viewpoints.
\textbf{(2) Geometry-aware encoding consistently outperforms absolute encoding.}
Both GTA and ViewRope outperform 3D RoPE, 
confirming the findings of prior work~\citep{li2025camerasrelativepositionalencoding,miyato2024gtageometryawareattentionmechanism} that relative geometric relationships are more effective than absolute coordinates.
\textbf{(3) ViewRope maintains competitive visual quality.}
Despite focusing on geometric consistency, ViewRope achieves comparable or better PSNR/SSIM than baselines, indicating that the geometric inductive bias does not sacrifice generation quality. 

\textbf{Comparison with State-of-the-Art Interactive World Models.}
We further compare ViewRope with two leading interactive world model systems: Matrix-Game-2~\citep{zhang2025matrixgameinteractiveworldfoundation} and HY-WorldPlay~\citep{li2025hunyuangamecrafthighdynamicinteractivegame}. ViewRope consistently outperforms both baselines across all evaluated rotation magnitudes (30\textdegree–75\textdegree).
The improvement is most pronounced in loop-closure error (LCE): ViewRope reduces LCE by 6.5\% compared to HY-WorldPlay at 30\textdegree, 7.9\% at 45\textdegree, and 11.4\% at 75\textdegree, demonstrating that per-patch geometric encoding provides stronger spatial memory than action-conditioned approaches.
Notably, the performance gap widens with increasing rotation angle, suggesting that ViewRope's ray-based attention becomes more beneficial for larger camera excursions where geometric correspondence is critical.
Qualitative comparisons and extended results including large-angle (90\textdegree--180\textdegree) trajectories are provided in Appendix~\ref{subsec:baseline_case_study}.

\subsection{Efficiency of Geometry-Aware Sparse Attention}
\label{subsec:efficiency}

\textbf{Experiments setup.}
To validate sparse attention, we continue training the models from Section~\ref{subsec:main} on 61-frame sequences with various sparse attention mechanisms (top-$k{=}5$) for 6k steps, then on 201-frame sequences for 2k steps to enable longer sequence generation. We evaluate on 90\textdegree\ and 180\textdegree\ scenarios which require longer sequences.

\begin{table*}[!t]
    \centering
    \caption{\textbf{Sparse attention comparison on ViewBench.} We report visual quality (PSNR$\uparrow$, SSIM$\uparrow$, LPIPS$\downarrow$) and geometric consistency (LCE$\downarrow$) for 90\textdegree\ and 180\textdegree\ view synthesis. Best in \textbf{bold}.}
    \small
    \label{tab:sparse_results}
    \vspace{-2mm}
    \begin{tabular}{c|c c c c|c c c c}
        \toprule
    \multirow{2}{*}{\textbf{Method}} & \multicolumn{4}{c|}{\textbf{90 deg}} & \multicolumn{4}{c}{\textbf{180 deg}} \\
     & \textbf{PSNR} & \textbf{SSIM} & \textbf{LPIPS $\downarrow$} & \textbf{LCE $\downarrow$} & \textbf{PSNR} & \textbf{SSIM} & \textbf{LPIPS $\downarrow$} & \textbf{LCE $\downarrow$} \\
    \midrule
        Sparse w/o ViewRope & 10.97 & 0.080  & 0.8887 & 0.8932 & 9.937 &  0.0618 & 0.9286 & 0.9243 \\
        GTA w/ Sparse & 8.603 & 0.0755 & 0.8316 & 0.8020 & 9.275 & 0.078 & 0.8062 & 0.7924 \\
        ViewRope w/ Sliding Window & 15.20 & 0.3701 & 0.5513 & 0.6543 & 14.44 & 0.3406 & 0.6139 & 0.6598 \\
        ViewRope w/ Sparse (Ours) & \textbf{15.61} & \textbf{0.4001} & \textbf{0.5382} & \textbf{0.5445} & \textbf{14.35} & \textbf{0.3458} & \textbf{0.6043} & \textbf{0.5609} \\
        \hline
        \end{tabular}
    \vspace{-3mm}
\end{table*}

Table~\ref{tab:sparse_results} shows the results of sparse attention comparison on ViewBench. We make two key observations:
\textbf{(1) ViewRope w/ Sparse consistently outperforms other methods}.
ViewRope w/ Sparse outperforms all baselines, reducing LCE by 16\% compared to sliding window attention. This demonstrates that our geometry-aware sparse attention mechanism is more effective for long-sequence view synthesis.
\textbf{(2) ViewRope stabilizes sparse training}.
We found that both na\"ive sparse attention (without geometric encoding) and GTA w/ Sparse suffer from loss divergence during training, whereas ViewRope w/ Sparse maintains stable convergence throughout. We attribute this stability to ViewRope's ray-based rotations, which impose geometrically meaningful structure on the Q/K dot products used for relevance scoring (Eq.~\ref{eq:block_affinity}). This structure yields more reliable frame selection and, consequently, more stable gradient signals during sparse training.

\textbf{Counterfactual Validation.}
To verify that our sparse selection is causally meaningful (not just random sparsity that happens to work), we conduct counterfactual experiments where we alter the selection strategy:
1) \textbf{Random Selection}: Select $k$ frames randomly from the entire history; 2)\textbf{Exclude Selected}: Explicitly exclude the top-$k$ frames identified by ViewRope, and randomly select $k$ frames from the remaining history.

Table~\ref{tab:counterfactual} shows that random selection causes 25.2\% LCE degradation. Crucially, when we explicitly exclude the frames identified as important by ViewRope (Exclude Selected), performance degrades even further by 38.1\%.
This confirms that our geometry-aware selection identifies the specific frames that are causally necessary for consistency.

\begin{table}[h]
    \centering
    \caption{\textbf{Counterfactual validation.} Explicitly excluding the frames selected by ViewRope causes the largest performance drop, confirming their causal importance.}
    \label{tab:counterfactual}
    \vspace{-2mm}
    \small
    \begin{tabular}{l|cc}
        \toprule
        \textbf{Condition} & \textbf{180 deg LCE}$\downarrow$ & \textbf{$\Delta$ LCE} \\
        \midrule
        Normal (Geo-Sparse) & 0.5609 & 0.0\% \\
        Random Selection & 0.7027 & +25.2\% \\
        Exclude Selected & 0.7744 & +38.1\% \\
        \bottomrule
    \end{tabular}
    \vspace{-3mm}
\end{table}

\begin{figure}[h]
    \centering
    \includegraphics[width=0.95\columnwidth]{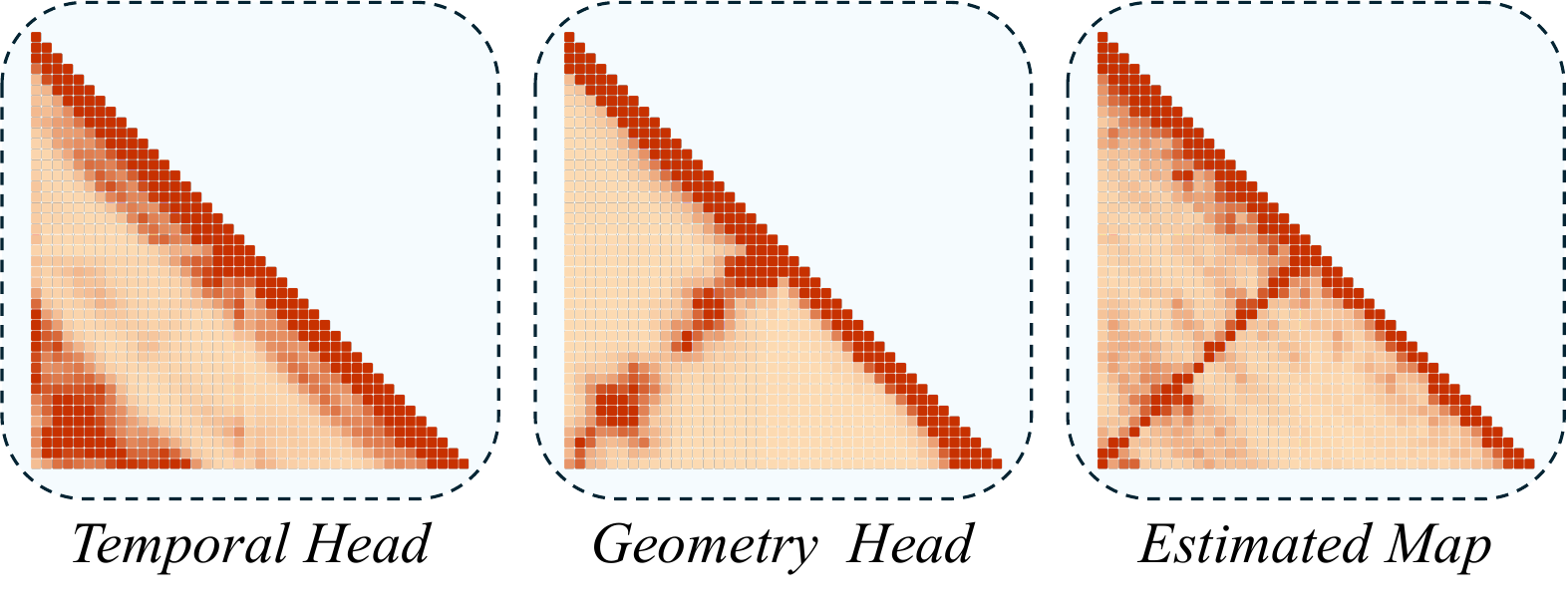}
    \vspace{-3mm}
    \caption{\textbf{Visualization of attention specialization.} Left: A standard temporal head focuses on recent or temporally periodic frames. Middle: A geometry-aware head captures long-range spatial overlap (evident in the antidiagonal activation during loop closure). Right: The aggregated attention map illustrates how geometric cues guide sparse block selection.}
    \label{fig:attnviz}
    \vspace{-3mm}
\end{figure}

\textbf{Visualizing Geometric Relationships.}
To further substantiate that the model learns meaningful geometric relationships, we visualize the attention maps for a loop closure sequence in Figure~\ref{fig:attnviz}.
We observe that different attention heads specialize in distinct patterns: while common heads (left) focus on temporal locality, specific geometry-sensitive heads (middle) emerge to capture spatial overlap.
Notably, these geometry heads exhibit high activation for temporally distant but spatially aligned frames.
This geometric signal is successfully integrated into the final estimated attention map (right), thereby guiding the block selection mechanism to retrieve the correct historical context.

\textbf{Efficiency Comparison.}
We evaluate computational efficiency. Sparse attention (top-$k=5$) reduces training time from 27.66 s/iter to 22.01 s/iter on 201-frame sequences, achieving a $\sim$25\% acceleration compared to dense attention.

\begin{figure*}[t]
    \centering
    \includegraphics[width=0.95\textwidth]{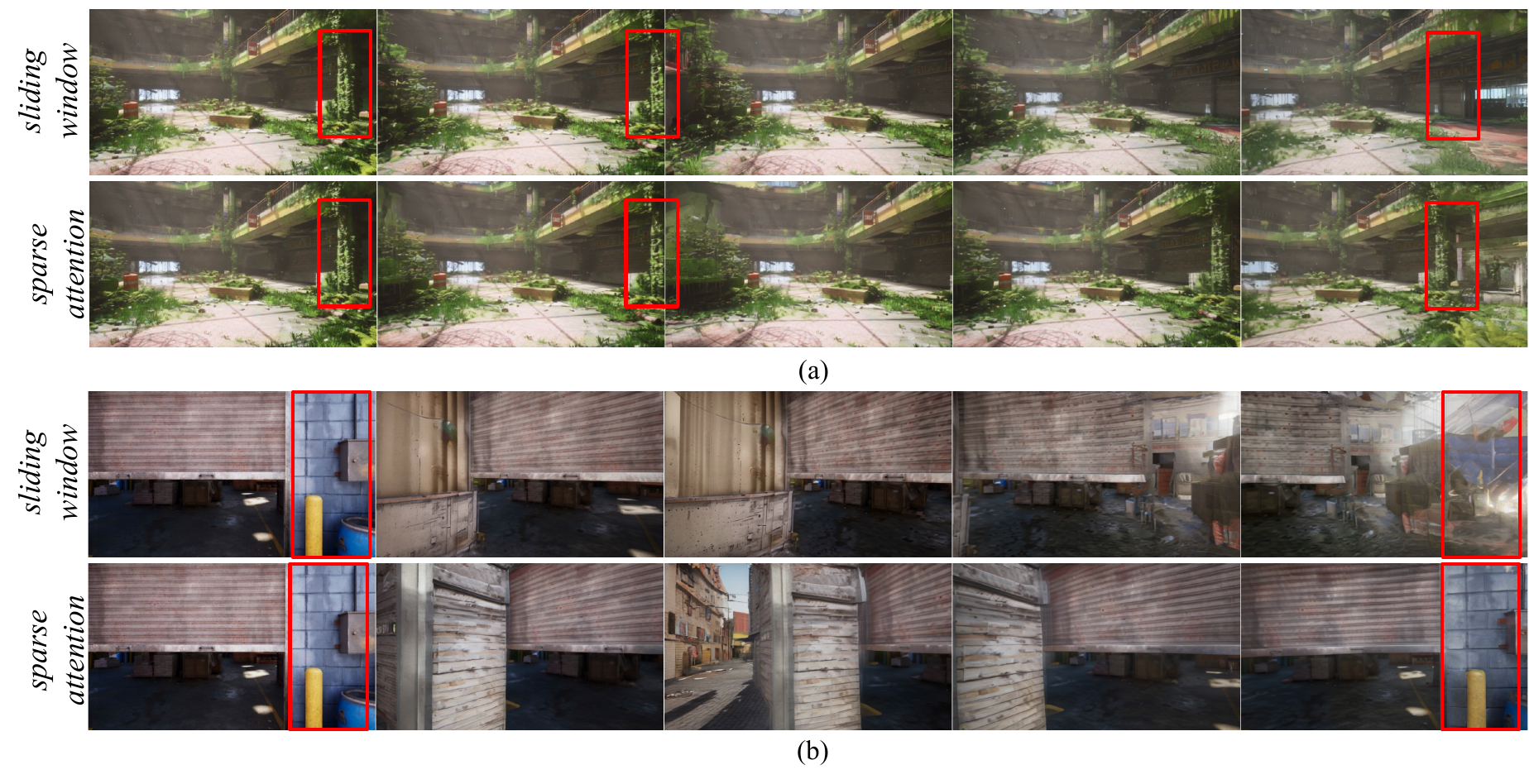}
    \vspace{-3mm}
    \caption{\textbf{Case study.} Upper and lower sequences show ViewRope with Sliding Window and Sparse attention, respectively.
    }
    \label{fig:cases_study}
    \vspace{-3mm}
\end{figure*}
\textbf{Case Study.}
Figure~\ref{fig:cases_study} shows the cases where viewrope w/ sparse outperforms other methods.
In Figure~\ref{fig:cases_study}(a), the green column on the right side disappears after the turn under the sliding window method, while ViewRope maintains it well.
In Figure~\ref{fig:cases_study}(b), the blue wall on the right side becomes blurry and exhibits noticeable drift and hallucinated details under the sliding window method, whereas ViewRope accurately recovers the original scene structure.
This demonstrates that ViewRope's geometry-aware sparse attention mechanism is more effective for long-sequence geometry-aware retrieval.

\subsection{Ablation Studies}
\label{subsec:ablation}

\textbf{Channel Allocation for ViewRope.}
\begin{table}[h]
\centering
\caption{\textbf{Ablation: ViewRope embedding strategies.} We compare different strategies for integrating ViewRope into the feature channels. Embedding in T-dimension low frequencies yields the best performance.}
\label{tab:ablation_channels}
\vspace{-2mm}
\small
\setlength{\tabcolsep}{3pt}
\begin{tabular}{l|c}
\toprule
\textbf{Embedding Strategy} & \textbf{Training Loss}$\downarrow$ \\
\midrule
T-dim low-freq (ch 32--44) & \textbf{0.0859} \\
H/W-dim low-freq (ch 74--86, 116--128) & 0.0861 \\
H/W-dim low-freq (replace 3D RoPE) & 0.0874 \\
All dimensions (ch 1--128) & 0.0894 \\
\bottomrule
\end{tabular}
\vspace{-3mm}
\end{table}
Efficiently integrating ViewRope into the existing 3D RoPE architecture is a key design challenge. The original model partitions RoPE into Temporal (T), Height (H), and Width (W) components, occupying 44, 42, and 42 dimensions respectively, totaling 128 dimensions.
We investigate four strategies for embedding ViewRope:
(1) Embedding in the lowest frequency bands of the T dimension (channels 32--44);
(2) Embedding in the lowest frequency bands of H and W dimensions (channels 74--86 and 116--128);
(3) Embedding in the lowest-frequency bands of H and W, while disabling the corresponding 3D RoPE components to prevent interference;
(4) Distributing ViewRope across all dimensions (1--128).
We evaluate these strategies using training loss as the metric.

Table~\ref{tab:ablation_channels} presents the ablation results for these embedding schemes.
Embedding in the lowest frequency bands of the T dimension yields the best performance (0.0859), suggesting that the temporal dimension offers the most suitable capacity for encoding view information without disrupting critical spatial feature representations governed by the H and W dimensions.
Interestingly, simply replacing the original 3D RoPE components with ViewRope leads to performance degradation (0.0874 vs. 0.0861), indicating that the original positional information remains complementary to our relative geometric encoding. 
Finally, distributing ViewRope across all dimensions results in the highest training loss (0.0894), likely due to the excessive interference with the backbone's pre-trained frequency structure.

\textbf{Number of Retrieved Frames.}
Due to space constraints, we provide the ablation study on the number of retrieved frames ($k$) in Appendix~\ref{subsec:topk_ablation}.
We observe that increasing $k$ generally improves visual quality (PSNR, SSIM) by accessing more texture details, but geometric consistency (LCE) peaks at the training setting ($k=5$), suggesting a trade-off between texture richness and geometric precision.

\section{Conclusion and Future Work}
\label{sec:conclusion}

In this work, we introduced \textbf{ViewRope}, a novel geometric positional encoding that effectively bridges the gap between 3D consistency and generative flexibility in video diffusion models. 
By treating ``view as position'' and embedding camera ray directions directly into the attention mechanism, our method enables the model to maintain long-term spatial persistence and geometric consistency, particularly in challenging loop-closure scenarios. 
Furthermore, our proposed \textbf{Geometry-Aware Sparse Attention} significantly improves computational efficiency by selectively attending to geometrically relevant historical frames, making long-video generation feasible without sacrificing performance. 
Experimental results on our newly proposed \textbf{ViewBench} demonstrate that our approach achieves state-of-the-art consistency while significantly reducing computational costs compared to dense attention baselines.

Despite these advancements, our method has limitations.
First, it may struggle with drastic scene transitions, such as moving from one room to another, where geometric correspondences between views are weak or nonexistent.
Integrating explicit 3D modeling with generative approaches and combining them with implicit representations remains to be explored. Further post-training with distillation and RL to support more dynamic scenarios and longer sequence generation remains as future research.

\section{Impact Statement}
\label{sec:impact}

This paper advances machine learning for interactive world models. 
Our view-consistent video generation could enhance VR/AR, entertainment, education, and training applications. 
However, high-quality video generation raises concerns about deepfake misuse and requires careful consideration of copyright and privacy protection in virtual content creation.
 We acknowledge the need for responsible development of such technologies.

\bibliographystyle{icml2026}
\bibliography{main}

\newpage
\appendix
\onecolumn

\section{Detailed Training And Inference Pipeline}
We provide the detailed training and inference pipeline with Frame Sparse Attention under Teacher Forcing as shown in Algorithm~\ref{alg:sparse_attn}.
\begin{algorithm}[h]
    \small
    \caption{Training and Inference with Frame Sparse Attention under Teacher Forcing}
    \label{alg:sparse_attn} 
    \begin{minipage}[t]{0.48\textwidth}
    \begin{algorithmic}[1]
    \STATE \underline{\textbf{Training Procedure:}}
    \STATE \textbf{Input:} Clean sequence $\mathbf{Z_0}$, Camera sequence $\mathcal{C}$, Text $\mathcal{Y}$
    \STATE \textbf{Noising:} Add noise $\mathbf{N}$ to $\mathbf{Z_0}$ to obtain $\mathbf{Z}_t$
    \STATE $\mathbf{pred} \gets \text{Model}(\mathbf{Z_0}, \mathbf{Z}_t, \mathcal{C}, \mathcal{Y})$
    \STATE $\mathcal{L} \gets \text{FlowMatching\_loss}(\mathbf{pred}, \mathbf{Z_0}, \mathbf{N})$
    \STATE \underline{\textit{Inside Attention:}}
    \STATE \quad \textbf{Input:} Clean $\mathbf{Q}_0$, $\mathbf{K}_0$, $\mathbf{V}_0$, Camera sequence $\mathcal{C}$, 
    \STATE \quad \textbf{Input:} Noise $\mathbf{Q}_t$, $\mathbf{K}_t$, $\mathbf{V}_t$.
    \STATE \quad \quad $\mathbf{Q}_0, \mathbf{Q}_t \gets \text{VR}(\mathbf{Q}_0, \mathcal{C}), \text{VR}(\mathbf{Q}_t, \mathcal{C})$
    \STATE \quad \quad $\mathbf{K}_0, \mathbf{K}_t \gets \text{VR}(\mathbf{K}_0, \mathcal{C}), \text{VR}(\mathbf{K}_t, \mathcal{C})$
    \STATE \quad \quad $\mathbf{Q}, \mathbf{K}, \mathbf{V} \gets [\mathbf{Q}_0; \mathbf{Q}_t], [\mathbf{K}_0; \mathbf{K}_t], [\mathbf{V}_0; \mathbf{V}_t]$
    \STATE \quad \quad Sample indices $\mathcal{I}$ randomly
    \STATE \quad \quad $\mathbf{s} \gets \mathbf{q}[\mathcal{I}] \cdot \mathbf{k}[\mathcal{I}]^\top$
    \STATE \quad \quad $\mathbf{s} \gets \text{ApplyTeacherForcingMask}(\mathbf{s})$
    \STATE \quad \quad $\mathbf{M} \gets \text{TopK}(\text{softmax}(\mathbf{s}))$
    \STATE \quad \quad $\mathbf{o} \gets \text{FrameSparseAttention}(\mathbf{q}, \mathbf{k}, \mathbf{v}, \mathbf{M})$
    \end{algorithmic}
    \end{minipage}
    \hfill
    \begin{minipage}[t]{0.48\textwidth}
    \begin{algorithmic}[1]
    \STATE \underline{\textbf{Inference Procedure:}}
    \STATE \textbf{Input:} First frame $\mathbf{x}_0$, Camera sequence $\mathcal{C}$, Text $\mathcal{Y}$
    \FOR{$c$ \textbf{in} $\mathcal{C}$}
        \STATE Initialize noise $\mathbf{n}_t$
        \FOR{$step$ \textbf{in} DenoiseSteps}
            \STATE $\mathbf{pred} \gets \text{Model}(\mathbf{n}_t, c, \mathcal{Y})$
            \STATE $\mathbf{n}_t \gets \text{ODE\_Update}(\mathbf{n}_t, \mathbf{pred})$
            \STATE \quad \textit{Inside Attention:}
            \STATE \quad \quad Given $\mathbf{q}_t$, $\mathbf{k}_0, \mathbf{v}_0$ (Clean KV Cache)
            \STATE \quad \quad Sample indices $\mathcal{I}$ randomly
            \STATE \quad \quad $\mathbf{q}_c, \mathbf{k}_c \gets \mathbf{q}_t[\mathcal{I}], \mathbf{k}_0[\mathcal{I}]$
            \STATE \quad \quad $\mathbf{s} \gets \mathbf{q}_c \cdot \mathbf{k}_c^\top$
            \STATE \quad \quad $\mathbf{M} \gets \text{TopK}(\text{softmax}(\mathbf{s}))$
            \STATE \quad \quad $\mathbf{o} \gets \text{FrameSparseAttention}(\mathbf{q}_t, \mathbf{k}_0, \mathbf{v}_0, \mathbf{M})$
        \ENDFOR
        \STATE $\mathbf{n}_0 \gets \mathbf{n}_t$
        \STATE $\text{Model}(\mathbf{n}_0, c, \text{cache=True})$ \COMMENT{Cache current frame}
    \ENDFOR
    \end{algorithmic}
    \end{minipage}
\end{algorithm}

\subsection{Training Configuration}
\label{app:implementation}

We provide detailed training configurations for reproducibility.

\paragraph{Model and Resolution.}
We build upon WAN 2.2 TI2V-5B~\citep{wan2025wanopenadvancedlargescale}, a text-and-image-to-video diffusion transformer with 5 billion parameters.
The training resolution is $480 \times 832$ with 61 frames per clip.
To convert the model into a streaming video generator, we use teacher-forcing training to align the model with the autoregressive generation interface and KV-caching behavior.

\paragraph{Optimization.}
We train with a batch size of 64 for 6k steps, using the AdamW optimizer with learning rate $5 \times 10^{-5}$ and linear warmup for 50 iterations.
The training process takes approximately 2 days on 16 NVIDIA A100 GPUs.

\paragraph{Training Data.}
The training data consists of three sources:
\begin{itemize}[leftmargin=*,itemsep=1pt,topsep=2pt]
    \item \textbf{Context-as-Memory}~\citep{yu2025contextmemorysceneconsistentinteractive}: ${\sim}$760k frames of yaw-only camera motion sequences.
    \item \textbf{GF-Minecraft}~\citep{yu2025gamefactorycreatingnewgames}: ${\sim}$4M frames of gameplay videos with diverse actions.
    \item \textbf{ViewBench}: ${\sim}$500k frames of synthetic sequences with complete 3-axis rotation coverage.
\end{itemize}
To obtain a balanced distribution of camera poses and scene geometries, we adjust the sampling rate of each dataset to achieve a 1:1:1 ratio during training.

\paragraph{Fair Comparison.}
All RoPE variants (3D RoPE, GTA, ViewRope) are applied to the same channels of the latent representation to ensure fair comparison.
The channel allocation follows the ablation study in Section~\ref{subsec:ablation}, where ViewRope is embedded in the lowest frequency bands of the temporal dimension (channels 32--44).

\section{Loop Closure Formulation}
\label{app:loop_closure_formulation}

This section provides the formal definition of loop-closure consistency referenced in Section~3.1.

Standard video generators enforce \emph{local temporal coherence} via objectives of the form
$\mathcal{L}_{\text{temp}}(\theta) := \mathbb{E}_{\mathbf{x} \sim p_\theta}\!\left[\sum_{t=2}^{T} d(\mathbf{x}_t,\mathbf{x}_{t-1})\right]$,
which only constrain adjacent frames and do not prevent long-horizon geometric drift.

A pose-conditioned world model must additionally satisfy loop closure. Define the revisit indicator
\begin{equation}
w_{t,k} := \mathbb{I}\!\left(\Delta(\mathcal{C}_t,\mathcal{C}_k)\le \varepsilon\right),\qquad k<t,
\end{equation}
where $\Delta(\cdot,\cdot)$ measures pose similarity and $\varepsilon$ is a tolerance threshold.

Let $\mathcal{W}_{k\leftarrow t}$ denote the image warp induced by the relative camera motion and scene geometry.
For a pixel $\mathbf{u}=(u,v)$ with homogeneous coordinate $\tilde{\mathbf{u}}=[u,v,1]^\top$, and a depth field $D_t$ at time $t$:
\begin{align}
\mathbf{X}_t(\mathbf{u}) &:= \mathbf{K}_t^{-1}\tilde{\mathbf{u}}\;D_t(\mathbf{u}),\\
\mathbf{X}_k(\mathbf{u}) &:= \mathbf{R}_k\mathbf{R}_t^{-1}\mathbf{X}_t(\mathbf{u})
+\big(\mathbf{P}_k-\mathbf{R}_k\mathbf{R}_t^{-1}\mathbf{P}_t\big),\\
\mathcal{W}_{k\leftarrow t}(\mathbf{u}) &:= \pi\!\big(\mathbf{K}_k\,\mathbf{X}_k(\mathbf{u})\big),
\end{align}
where $\pi(\cdot)$ denotes perspective division. The loop closure loss $\mathcal{L}_{\text{lc}}(\theta)$ enforces $\mathbf{x}_t(\mathbf{u}) \approx \mathbf{x}_k(\mathcal{W}_{k\leftarrow t}(\mathbf{u}))$ for $\mathbf{u}\in\Omega_{t,k}$ (the mutually visible region). We formalize this as:
\begin{equation}
\mathcal{L}_{\text{lc}}(\theta)
:=
\mathbb{E}_{\mathbf{x} \sim p_\theta}\!\left[
\sum_{t=1}^{T}\sum_{k<t} w_{t,k}
\sum_{\mathbf{u}\in\Omega_{t,k}}
\rho\!\left(
\mathbf{x}_t(\mathbf{u}) -
\mathbf{x}_k\!\left(\mathcal{W}_{k\leftarrow t}(\mathbf{u})\right)
\right)
\right],
\end{equation}
where $\rho(\cdot)$ is a robust penalty (e.g., Huber).

\section{ViewBench Benchmark}
\label{app:viewbench}

This appendix provides additional details on ViewBench. For the main comparison with existing datasets, see Table~\ref{tab:viewbench_comparison} in Section~\ref{subsec:setup}. For baseline comparisons with Matrix-Game-2 and HY-WorldPlay, see Table~\ref{tab:viewbench_baselines_full} and the accompanying analysis.

\subsection{Extended Dataset Comparison}
\label{app:viewbench_overview}

Table~\ref{tab:viewbench_comparison_full} provides an extended comparison with additional properties beyond those shown in the main text.

\begin{table}[h]
\centering
\caption{\textbf{Extended comparison with existing datasets.}}
\label{tab:viewbench_comparison_full}
\vspace{-2mm}
\small
\begin{tabular}{l|ccc}
\toprule
\textbf{Property} & \textbf{CaM} & \textbf{GF-MC} & \textbf{ViewBench} \\
\midrule
Yaw & $\checkmark$ & $\checkmark$ & $\checkmark$ \\
Pitch & $\times$ & $\checkmark$ & $\checkmark$ \\
Roll & $\times$ & $\times$ & $\checkmark$ \\
Loop-closure trajectories & $\times$ & $\times$ & $\checkmark$ \\
Controlled angle magnitudes & $\times$ & $\times$ & $\checkmark$ \\
Per-frame SE(3) c2w & partial & $\times$ & $\checkmark$ \\
Overlap annotations & FOV-based & $\times$ & depth-based \\
Engine & UE5  & MC & UE5 \\
\bottomrule
\end{tabular}
\end{table}

\subsection{Scene Environments}
\label{app:viewbench_scenes}

ViewBench comprises 10 photorealistic UE5 environments spanning indoor, outdoor, urban, and natural settings (Table~\ref{tab:viewbench_scenes}). The diversity in geometry, lighting, and texture ensures evaluation results generalize across visual conditions.

\begin{table}[h]
\centering
\caption{\textbf{ViewBench scenes.}}
\label{tab:viewbench_scenes}
\vspace{-2mm}
\small
\begin{tabular}{l|l|l}
\toprule
\textbf{Scene} & \textbf{Type} & \textbf{Description} \\
\midrule
Abandoned\_HongKong & Outdoor/Urban & Mid-scale urban ruins \\
Abandoned\_Mall & Indoor & Two-floor shopping mall \\
DeadCity & Outdoor/Urban & Derelict city, dark lighting \\
PostApocalypticCity & Outdoor/Urban & Large-scale post-apocalyptic \\
FPS\_Template & Outdoor/Desert & Middle-Eastern battlefield \\
Container\_Yard & Outdoor/Industrial & Container stacking yard \\
Rome & Outdoor/Historical & Roman-style architecture \\
SuburbsCityPack & Outdoor/Suburban & Suburban street scene \\
ChineseAlley & Outdoor/Cultural & Chinese-style alleyway \\
UrbanDistrict\_Gate & Indoor-like & Narrow shantytown alleys \\
\bottomrule
\end{tabular}
\end{table}

\subsection{Camera Rotation Ranges}
\label{app:viewbench_rotation}

ViewBench trajectories are organized into two parts.

\paragraph{Part 1: Pure rotation.}
The camera is stationary and performs rotation-only motion covering all 7 axis combinations (3 single-axis + 3 dual-axis + 1 triple-axis). Each trajectory follows a \textit{rotate-away--rotate-back} loop-closure design, with rotation magnitudes sampled from \{30\textdegree, 75\textdegree, 90\textdegree, 180\textdegree\}. For each of 10 scenes $\times$ 7 axis combinations, we generate 100 samples, yielding ${\sim}$7{,}000 clips.

\paragraph{Part 2: Rotation + translation.}
The camera moves within a compact exploration radius while simultaneously rotating, mimicking interactive navigation.
Each sequence is composed of actions randomly drawn from four types:
\textbf{RotateOnly} (in-place camera rotation),
\textbf{MoveOnly} (pure WASD translation without rotation),
\textbf{MoveAndRotate} (simultaneous translation and rotation),
and \textbf{Orbit} (circular motion around a point of interest).
All action types may include roll rotation with type-specific probabilities and ranges.

\subsection{Data Format}
\label{app:viewbench_format}

Each frame is annotated with a $4\times4$ camera-to-world (c2w) SE(3) matrix, Euler angles (pitch, roll, yaw), position in centimeters, FOV (including the vertical one and the horizontal one), and binary WASD key states. The rotation matrix follows a ZYX convention ($\mathbf{R} = \mathbf{R}_z\text{(yaw)} \cdot \mathbf{R}_y\text{(pitch)} \cdot \mathbf{R}_x\text{(roll)}$) in UE5's left-handed coordinate system (X-Forward, Y-Right, and Z-Up). A post-processing pipeline converts ViewBench data into the action formats used by CaM and GF-Minecraft, enabling unified evaluation. Depth-based frame overlap annotations are also provided for attention recall analysis.

\subsection{Training and Evaluation Splits}
\label{app:viewbench_splits}

\paragraph{Training.}
The training set combines Part~1 (pure rotation) and Part~2 (rotation + translation) data, totaling 1,059 video sequences (${\sim}$500k frames at 30\,fps) across 10 scenes. During training, ViewBench data is mixed with CaM and GF-Minecraft at a 1:1:1 sampling ratio. The evaluation set is \emph{separately collected} from the training set with non-overlapping trajectories, ensuring no data leakage between training and evaluation.

\paragraph{Evaluation.}
The evaluation set consists of separately collected Part~1 pure-rotation loop-closure trajectories from the same 10 scenes, totaling 600 samples. Each sample is downsampled to 16\,fps and contains 61 frames, providing a first frame, the full camera trajectory with per-frame SE(3) poses, and ground-truth UE5-rendered video.
We evaluate frame-level PSNR, SSIM, and LPIPS against the ground truth, as well as Loop Closure Error (LCE):
\begin{equation}
    \text{LCE} = \text{LPIPS}(\mathbf{x}_0,\, \hat{\mathbf{x}}_T),
\end{equation}
where $\mathbf{x}_0$ is the ground-truth first frame and $\hat{\mathbf{x}}_T$ is the generated frame at the return pose. LCE isolates the challenge of remembering previously seen content after an extended camera excursion.

\subsection{Complete Baseline Results on ViewBench}
\label{app:viewbench_baselines}

Table~\ref{tab:viewbench_baselines_full} in the main text presents results for small-to-medium angles (30\textdegree, 45\textdegree, 75\textdegree) where ViewRope consistently outperforms all baselines. Here we provide the complete results including large-angle (90\textdegree, 180\textdegree) trajectories.

We evaluate representative baselines that accept a first frame and per-frame action sequence. For fair comparison, baseline evaluation uses only yaw/pitch trajectories (no roll). ViewRope is additionally evaluated on the full set including roll. The unified evaluation script automatically converts ViewBench SE(3) poses to each model's expected action format (e.g., mouse deltas for Matrix-Game-2, yaw-only deltas for CaM).

\begin{table}[h]
\centering
\caption{\textbf{Complete baseline results on ViewBench} across all rotation magnitudes. Best in \textbf{bold}. See Section~\ref{subsec:main} for analysis.}
\label{tab:viewbench_baselines_full}
\vspace{-2mm}
\small
\setlength{\tabcolsep}{4pt}
\begin{tabular}{cl|cccc}
\toprule
\textbf{Angle} & \textbf{Model} & PSNR$\uparrow$ & SSIM$\uparrow$ & LPIPS$\downarrow$ & LCE$\downarrow$ \\
\midrule
\multirow{3}{*}{30\textdegree}
& Matrix-Game-2 & 14.27 & 0.2806 & 0.5723 & 0.5553 \\
& HY-WorldPlay & 17.04 & 0.4238 & 0.4697 & 0.4811 \\
& ViewRope (ours) & \textbf{17.53} & \textbf{0.4378} & \textbf{0.4080} & \textbf{0.4497} \\
\midrule
\multirow{3}{*}{45\textdegree}
& Matrix-Game-2 & 13.55 & 0.2535 & 0.6071 & 0.6175 \\
& HY-WorldPlay & 16.38 & 0.4085 & 0.5096 & 0.4944 \\
& ViewRope (ours) & \textbf{16.74} & \textbf{0.4130} & \textbf{0.4527} & \textbf{0.4545} \\
\midrule
\multirow{3}{*}{75\textdegree}
& Matrix-Game-2 & 13.46 & 0.2625 & 0.6084 & 0.6288 \\
& HY-WorldPlay & 15.19 & 0.3847 & 0.5643 & 0.5151 \\
& ViewRope (ours) & \textbf{15.27} & \textbf{0.3916} & \textbf{0.5398} & \textbf{0.4562} \\
\midrule
\multirow{3}{*}{90\textdegree}
& Matrix-Game-2 & 12.41 & 0.1878 & 0.6684 & 0.7121 \\
& HY-WorldPlay & \textbf{16.56} & \textbf{0.4174} & \textbf{0.4970} & \textbf{0.4169} \\
& ViewRope (ours) & 15.61 & 0.4001 & 0.5382 & 0.5445 \\
\midrule
\multirow{3}{*}{180\textdegree}
& Matrix-Game-2 & 12.74 & 0.2033 & 0.6310 & 0.6732 \\
& HY-WorldPlay & \textbf{14.82} & 0.3403 & \textbf{0.5978} & \textbf{0.4413} \\
& ViewRope (ours) & 14.35 & \textbf{0.3458} & 0.6043 & 0.5609 \\
\bottomrule
\end{tabular}
\end{table}

\paragraph{Extended analysis of large-angle performance.}
At large angles (90\textdegree--180\textdegree), ViewRope shows lower performance than HY-WorldPlay. We identify two factors \emph{orthogonal} to our positional encoding contribution:

\textbf{(1) Evaluation frame-rate mismatch.}
To fit large-angle round-trip trajectories into a fixed length of 161 frames, the evaluation sequences are uniformly resampled with endpoints held fixed, resulting in per-frame angular steps that substantially exceed the constant angular velocity seen during training. As a consequence, our model under-rotates (e.g., achieving ${\sim}$80\textdegree\ when 180\textdegree\ is requested), producing a large LCE at the return pose. HY-WorldPlay is trained on data with variable frame rates and employs Dual Action conditioning with RL-based post-training (WorldCompass) to explicitly optimize action following, making it more robust to such speed variation.

\textbf{(2) Error accumulation in teacher-forcing models.}
Our model is trained with teacher forcing, where the ground-truth context is provided during training. At inference time, each frame is autoregressively conditioned on previously \emph{generated} frames, so errors compound over the longer sequences required by large-angle trajectories. HY-WorldPlay mitigates this through Context Forcing distillation and 4-step denoising with self-correction, substantially reducing error accumulation.

These limitations are system-level and can be addressed independently of ViewRope. Combining ViewRope with advanced training strategies such as self-forcing~\citep{huang2025selfforcingbridgingtraintest} and RL-based action post-training is a promising direction for future work.

\subsection{Qualitative Comparison}
\label{subsec:baseline_case_study}

We present qualitative comparisons between Matrix-Game-2.0 (M-G~2.0), HY-WorldPlay (HY-World), and ViewRope (Ours) on ViewBench loop-closure trajectories. Each case shows the input first frame (left) followed by keyframes sampled from the generated video. Arrow icons indicate the camera rotation direction at each keyframe. The camera first rotates away from the starting viewpoint and then returns, forming a closed loop.

\begin{figure*}[h]
    \centering
    \includegraphics[width=\textwidth]{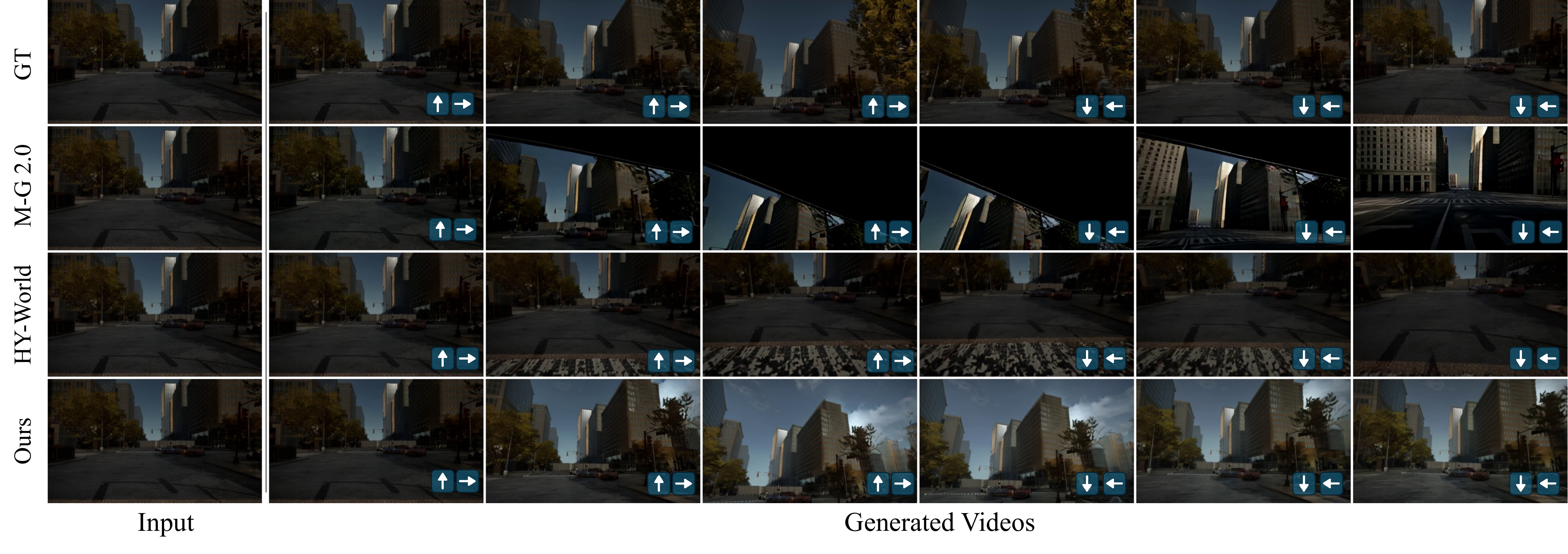}
    \caption{\textbf{Case 1: Yaw + Pitch loop closure in an urban street.} M-G~2.0 suffers from brightness collapse. HY-WorldPlay exhibits geometric drift. ViewRope maintains structural and lighting consistency.}
    \label{fig:case_study_1}
\end{figure*}

\paragraph{Case 1: Yaw + Pitch in an urban street scene (Figure~\ref{fig:case_study_1}).}
The camera rotates upward and rightward, then reverses back to the starting view. M-G~2.0 produces severe brightness collapse mid-trajectory, losing nearly all scene content in the dark frames. HY-WorldPlay maintains plausible appearance but exhibits geometric drift---building structures shift position upon return. ViewRope preserves both the scene structure and lighting conditions throughout the trajectory, yielding a return frame closely matching the ground truth.

\begin{figure*}[h]
    \centering
    \includegraphics[width=\textwidth]{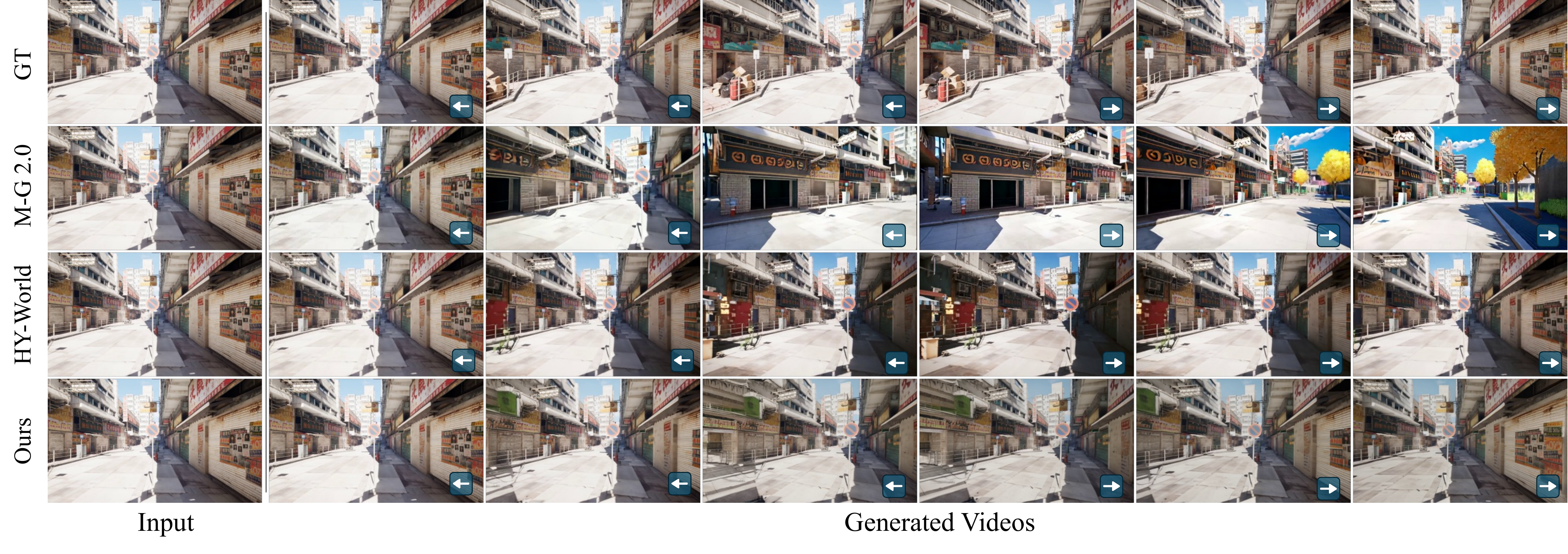}
    \caption{\textbf{Case 2: Pure yaw loop closure in an Asian street.} M-G~2.0 hallucinates entirely different content on return. HY-WorldPlay introduces nonexistent elements. ViewRope recovers the original scene faithfully.}
    \label{fig:case_study_2}
\end{figure*}

\paragraph{Case 2: Pure yaw in an Asian street scene (Figure~\ref{fig:case_study_2}).}
The camera pans left and then reverses rightward to return. M-G~2.0 generates quite big hallucination, introducing new scene elements (e.g., yellow trees) that are absent in the ground truth upon return. In this case, both HY-WorldPlay and ViewRope accurately recover the original storefronts and street layout, demonstrating strong long-term spatial memory.

\begin{figure*}[h]
    \centering
    \includegraphics[width=\textwidth]{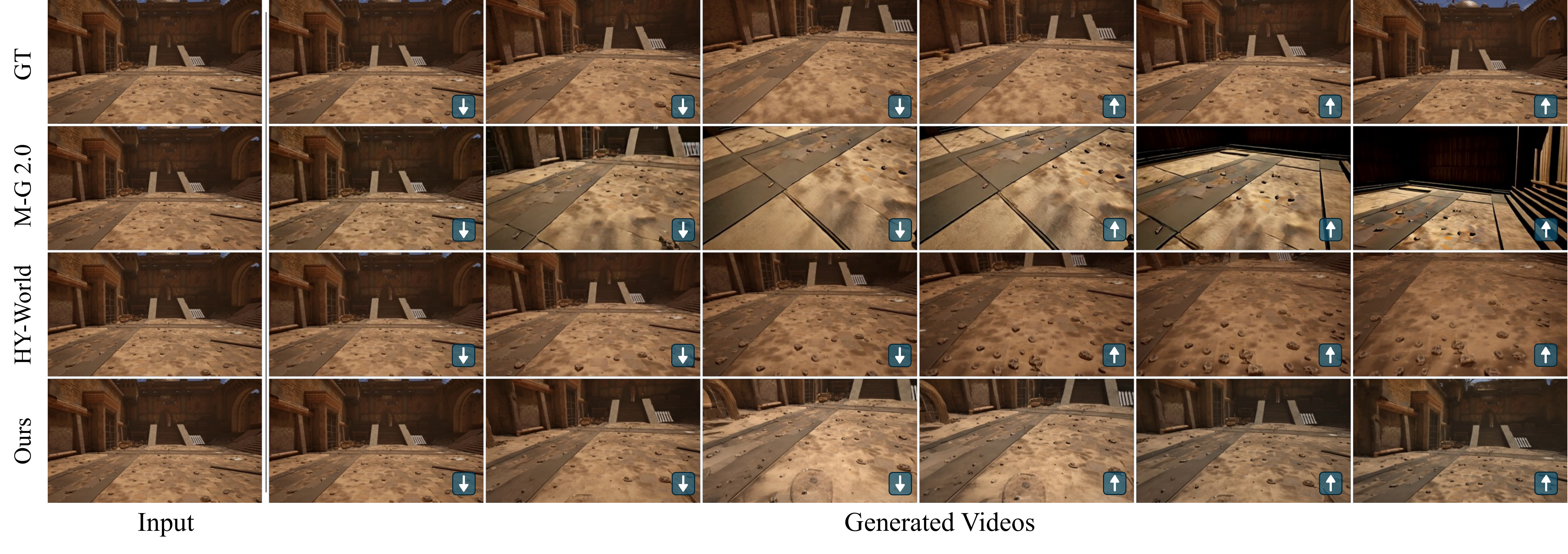}
    \caption{\textbf{Case 3: Pure pitch loop closure in Roman architecture.} M-G~2.0 generates a completely different scene upon return. HY-WorldPlay produces blurry, inconsistent structures. ViewRope accurately restores the original arched architecture.}
    \label{fig:case_study_3}
\end{figure*}

\paragraph{Case 3: Pure pitch in a Roman architecture scene (Figure~\ref{fig:case_study_3}).}
The camera tilts downward toward the ground and then pitches back up to the starting view. This tests vertical rotation consistency. M-G~2.0 fails catastrophically on the return---the final frame shows a completely different indoor scene with wooden structures, indicating total loss of scene identity. HY-WorldPlay maintains the general scene category but produces blurry architecture and seemingly fails to return. 
Comparatively, ViewRope faithfully reproduces the arched stone structures visible in the starting frame.

\newpage
\section{Additional Results}

\subsection{Ablation of Number of Topk Frames}
\label{subsec:topk_ablation}

As shown in Table~\ref{tab:topk_ablation}, we report the ablation results of the number of topk frames on ViewBench.
We use the model trained with topk=5 and adjust the number of retrieved frames at inference time to 1, 3, 10, and 20.
Increasing the number of retrieved frames generally improves visual quality metrics (PSNR, SSIM, and LPIPS), suggesting that accessing more reference frames provides richer texture details for generation.
However, we observe that the Loop Closure Error (LCE) achieves its optimum at topk=5 and degrades as the number of frames increases further.
This indicates that while more context helps visual quality, the model, having been trained with topk=5, may be distracted by the additional retrieved frames or struggle to effectively utilize the expanded context for long-term geometric consistency.

\begin{table*}[h]
    \centering
    \caption{\textbf{Ablation of number of topk frames on ViewBench.} We report visual quality (PSNR$\uparrow$, SSIM$\uparrow$, LPIPS$\downarrow$) and geometric consistency (LCE$\downarrow$) for 90\textdegree\ and 180\textdegree\ view synthesis. Best in \textbf{bold}.}
    \label{tab:topk_ablation}
    \small
    \vspace{-2mm}
    \begin{tabular}{c|c c c c|c c c c}
        \hline
         & \multicolumn{4}{c|}{\textbf{90 deg}} & \multicolumn{4}{c}{\textbf{180 deg}} \\
        \hline
         & \textbf{PSNR} & \textbf{SSIM} & \textbf{LPIPS $\downarrow$} & \textbf{LCE $\downarrow$} & \textbf{PSNR} & \textbf{SSIM} & \textbf{LPIPS $\downarrow$} & \textbf{LCE $\downarrow$} \\
         \midrule
        Top 1 & 13.01 & 0.3328 & 0.8230 & 0.8416 & 13.00 & 0.3337 & 0.8357 & 0.8109 \\
        Top 3 & 15.11 & 0.3792 & 0.5470 & 0.5777 & 14.54 & 0.3559 & 0.6096 & 0.5413 \\
        Top 5 & 15.71 & 0.3929 & 0.5267 & \textbf{0.5591} & 14.92 & 0.3574 & 0.5991 & \textbf{0.5385} \\
        Top 10 & 15.53 & 0.3951 & 0.5386 & 0.5804 & \textbf{14.99} & 0.3548 & 0.5997 & 0.5956 \\
        Top 20 & \textbf{16.17} & \textbf{0.4087} & \textbf{0.5240} & 0.5860 & 14.92 & \textbf{0.3576} & \textbf{0.5976} & 0.5976 \\
        \hline
        \end{tabular}
    \vspace{-3mm}
\end{table*}

\end{document}